\documentclass{article}

\usepackage[preprint,nonatbib]{tackling_climate_workshop_style}

\usepackage[utf8]{inputenc} % allow utf-8 input
\usepackage[T1]{fontenc}    % use 8-bit T1 fonts
\usepackage{hyperref}       % hyperlinks
\usepackage{url}            % simple URL typesetting
\usepackage{booktabs}       % professional-quality tables
\usepackage{amsfonts}       % blackboard math symbols
\usepackage{nicefrac}       % compact symbols for 1/2, etc.
\usepackage{microtype}      % microtypography
\usepackage{multirow}       % Added to format the tables
\usepackage{float}
\usepackage{graphicx}
\usepackage[backend=biber, style=numeric]{biblatex}

\title{Comparing the carbon costs and benefits of low-resource solar nowcasting\\
}

\author{%
  Ben Dixon\thanks{Primary author}\\
  University College London (UK)\\
  \texttt{benjamin.dixon.21@ucl.ac.uk} \\
  \And
  Jacob Bieker\\
  Open Climate Fix\\
  \texttt{jacob@openclimatefix.org} \\
    \And
  Mar\'ia P\'erez-Ortiz \\
  AI Centre, University College London (UK)\\
  \texttt{maria.perez@ucl.ac.uk} \\
}

\begin{document}

\maketitle

\begin{abstract}
Mitigating emissions in line with climate goals requires the rapid integration of low carbon energy sources, such as solar photovoltaics (PV) into the electricity grid. However, the energy produced from solar PV fluctuates due to clouds obscuring the sun's energy. Solar PV yield nowcasting is used to help anticipate peaks and troughs in demand to support grid integration. This paper compares multiple low-resource approaches to nowcasting solar PV yield.
To do so, we use a dataset of UK satellite imagery and solar PV energy readings over a 1 to 4-hour time range. Our work investigates the performance of multiple nowcasting models.
The paper also estimates the carbon emissions generated and averted by deploying models such as these, and finds that short-term PV forecasting may have a benefit several orders of magnitude greater than its carbon cost and that this benefit holds for small models that could be deployable in low-resource settings around the globe.  \end{abstract}

 %and estimates the carbon emissions generated by and averted through PV yield forecasting. We find that convolutional recurrent models tend to perform slightly better than purely convolutional models, most significantly when taking satellite images as input. %By exploring similar the activations for a similar CNN, we find that the model may be optimising for an undesireable outcome, in this case recognising contrast rather than learning cloud motion. 
%assessing whether explicitly representing the task as a recurrent time series leads to greater predictive accuracy. 

%%%%%%%%%%%%%%%%%%%%% INTRODUCTION %%%%%%%%%%%%%%%%%%%%%%%%%%%%%%%%%%

\section{Introduction}
Solar photovoltaics (PV) is one of the safest and cleanest sources of energy, as measured in death rates from accidents and air pollution, and greenhouse gas emission \cite{ritchie2021safest}. However, the fluctuations in the energy output, or yield, from solar PV makes it challenging for the electricity grid to rely at large scale on solar energy. The yield of solar PV is inherently uncertain because clouds can obscure the sun, and clouds can become thicker, thinner, and their movement across the sky is hard to predict \cite{boucher-clouds}. In order to guarantee stable energy supply, grid operators often have to keep gas spinning-reserve online, releasing carbon dioxide into the atmosphere \cite{GAO201535}. Forecasting over less than one day in advance is called `nowcasting' \cite{ConvLSTM} to emphasize its dynamic use. Nowcasting solar energy input into the grid accurately could reduce the need to keep the reserves on standby, allowing to use every kilowatt efficiently and enhancing the grid resilience to the fluctuations of renewable energy sources.

Since 2019,
Open Climate Fix (OCF) has been working with National Grid ESO, the electricity system operator for Great Britain, to build accurate PV nowcasting models that can enhance the resilience of the energy grid \cite{natgrideso_road_zero_carbon}. This paper complements the work done by OCF. Specifically, we research alternative low-resource methods (in terms of smaller/shallower neural networks architectures) for predicting PV yield, and analyse their performance, their carbon footprints and aim to estimate, even if roughly, the net emissions that could be averted by their potential use.

As more of the world incorporates solar PV into energy grids, there will be an increased need to develop accurate solar PV forecasts.   While a body of recent work has emphasized building more computationally efficient models, most of the machine learning work still focuses on the opposite: building larger
models with more parameters to tackle more complex tasks \cite{systematic-reporting-footprints}. Recent work poses the question of how much is the performance gain worth \cite{systematic-reporting-footprints}, specially in these applications where deployment of a simple technology could already have a big impact. The move towards less computationally intensive models has already happened once in weather forecasting, as deep learning, despite its competitiveness, has often lower computational requirements than numerical weather prediction based on differential equations \cite{royal-soc-can-beat, ConvLSTM}.
 
Through our work, we study low-resource and shallower predictive models and try to estimate the benefit that they could bring if they were deployed large-scale for solar nowcasting\footnote{Our low-resource settings are not simulated, this project has been done as part of a MSc thesis dissertation, with no other computational resources.}. The models, which are build on one  Tesla T4 GPU available through Google Colab, could be deployed large-scale around the world independently of hardware limitations. The reason we are interested in researching this low-resource setting, by studying smaller models, is because those are the models that could be deployed to support the grid in emerging economies, which are where the real need is for clean energy solutions, i.e. where the energy demand is rapidly growing.

%%%%%%%%%%%%%%%%%%%%% EXPERIMENTS %%%%%%%%%%%%%%%%%%%%%%%%%%%%%%%%%%

\section{An empirical comparison of PV yield nowcasting: Experimental results}
%%%%%%%%%%%%%%%%%%%%% DATA SOURCES %%%%%%%%%%%%%%%%%%%%%%%%%%%%%%%%%%

For the task at hand, we study convolutional neural networks (CNNs), Long Short-Term Memory networks (LSTMs) and their intersection \cite{ConvLSTM} (ConvLSTMs).
%The ConvLSTM was introduced by Shi et al in 2015 \cite{ConvLSTM} combining ideas from convolutional neural networks (CNNs) and Long Short-Term Memory networks (LSTMs). The authors demonstrate that it outperforms the state-of-the-art NWP optical flow algorithm in predicting precipitation over a 90 minute window. 
In theory, this comparison may seem futile as LSTMs and ConvLSTMs should outperform CNNs, because of their use of state information. However, Bai et al \cite{cnns-vs-lstms} argue that simpler CNNs can often outperform LSTMs. Moreover, ConvLSTMs require a large amount of memory simultaneously available in order to process and update the state information used in its predictions, which means they often take longer to train, leading to more energy usage and hence higher  emissions. 

\paragraph{Dataset}Solar generation data refers to the yield from specific solar panels.
The dataset\footnote{Thanks to Open Climate Fix for providing the datasets.} is composed of satellite imaging and PV readings. The satellite images are provided by EUMETSAT, covering the northern third of the Meteosat disc every five minutes. Open Climate Fix developed the \verb|eumetsat_uk_hrv| dataset \cite{eumetsat_dataset} which takes the `high resolution visible' (HRV) channel for a reduced time period and geospatial extent. The dataset was reduced in this paper to meet computing resources available. A 64 x 64 crop was randomly chosen to focus on Devon, selecting images taken between 05:00 and 20:00 as there is minimal PV output outside of this window for most of the year. However, this still left many readings in winter when the sun was below the horizon, and the PV yield was zero, so readings when the solar altitude was below 10 degrees were also dropped. 
 Open Climate Fix provided a dataset of 1311 energy yield readings from PV systems in 5 minute increments for most of 2018-2021, originally provided by a PV operator. The size of the dataset varied slightly based on the prediction window. For the shortest window, there were 2018 observations of (12, 64, 64) and (12, 1) to predict a series of 12 readings. For the longest prediction window, there were 659 blocks of the same sizes described earlier to predict a series of 48 readings.  

\paragraph{Learning task}
We formulate our learning task as predicting a sequence of values representing the future PV yield. All models take as input 12 sets of 5 min data, i.e., 1 hour of data as input, and predict forward between 1 and 4 hours. We also consider two learning scenarios: i) learning exclusively from past PV data only, and ii) learning from both past PV and satellite data as inputs.
 
\begin{figure}[ht]
\begin{center}
\includegraphics[width=9cm]{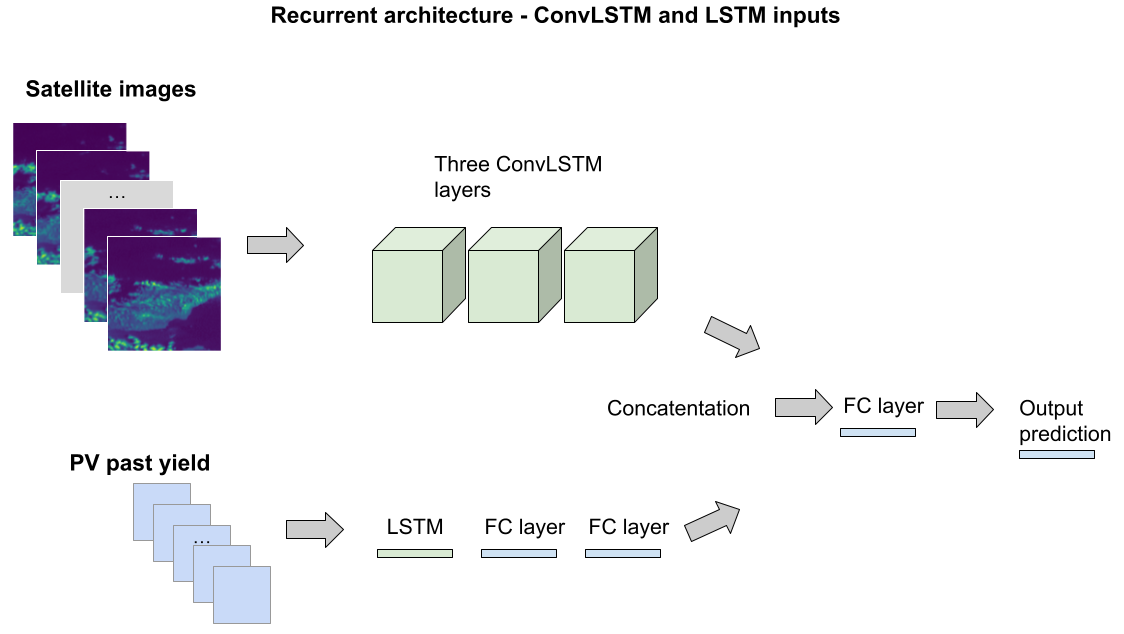}\label{convlstm_structure}
\caption{The ConvLSTM model, using both satellite images and past PV yield.}
\end{center}
\end{figure}

 \paragraph{Models compared}
For each learning scenario (with and without satellite imaging), we test both CNNs and LSTMs, our objective being comparing their prediction efficiency and carbon footprint at the task at hand. Additionally, LSTMs encode time relationships explicitly, and we are interested in evaluating if this additional complexity increases the prediction performance of our models to a significant extent (and at what potential carbon cost).
For the PV only models (CNN and LSTM), which are the simplest models we test, we take in past PV through multiple layers of CNNs, followed by a fully connected layer. The other PV only model has a similar structure but uses LSTMs in place of CNNs.  
The second group of models take both satellite images and past PV yield as input. The Conv3D is a CNN which uses multiple levels to try to learn abstract features. After each convolutional layer, a MaxPooling layer is used to reduce the dimensionality and learn the features. This is followed by two dense layers to resize the output to predict a sequence of values. The CNN treats the sequence as one vector. The ConvLSTM model (see \ref{convlstm_structure}) is similar to the Conv3D but uses recurrent modelling, with ConvLSTMs to process the images and LSTMs to process the PV input sequence. 

\paragraph{Model training and dataset split}
The model is trained on 22 months of data from 1/1/2020 to 7/11/2021. To avoid autocorrelation, the datasets are drawn from each month: days 1 to 20 are for training, days 22 to 24 for validation, and days 27 to 29 for testing. For all models, the models are trained to optimise MSE until convergence. Bayesian hyperparameter tuning was done:  The learning rate was varied between [0.01, 0.001, 0.0001, 0.00001], and dropout rate was varied between [0.05, 0.5] in steps of 0.05. For the CNNs and the ConvLSTMs, the kernel size was varied between [3,5,7]. For ConvLSTMs the number of filters was varied between [16,32,64]. For the fully connected layers the number of nodes was varied in the set [12,24,48]. The ConvLSTM has three ConvLSTM layers applied to images, and one LSTM layer applied to past PV.  The CNN model has three convolutional layers applied to the images, and one convolutional layer applied to past PV readings. For the PV-only models, the LSTM model has three layers of LSTMs, and the CNN model has three layers of CNNs.

\subsection{Results}
The task is to predict a sequence of values of variable lengths, from 1 hour to 4h. Table \ref{accuracy-table} shows normalised MAE and RMSE by model over different prediction windows. All models outperform the persistence baseline, often used for physical processes. ConvLSTM outperforms other models, especially over longer time horizons. Interestingly, an LSTM model taking PV only delivers results competitive with a Conv3D which also uses satellite data five out of eight times. Does time series modelling improves predictions? The recurrent designs have a lower error metric five out of eight times, but often by less than 0.2\%. Do satellite images improve performance? Yes, on six of eight times, and by margins approximately increasing 0.2\%-1.2\%.  
Overall, it appears that explicit modelling of temporal information improves the performance to a large extent. This is most significant for the comparison between ConvLSTM and CNN, which is when the inputs include image data. Perhaps this is because image data carries important trend information, such as motion of clouds.

\begin{table}[h!]
\label{accuracy-table}
\scriptsize
\centering
\setlength{\tabcolsep}{12pt}
\renewcommand{\arraystretch}{1.1}
\begin{tabular}{c c  c  c c  c c}
    & & Baseline & \multicolumn{2}{c}{PV only} & \multicolumn{2}{c}{Sat and PV}  \\ \hline
	Time & Metric & Persistence & CNN & LSTM & Conv3D & ConvLSTM \\
    \hline
	\multirow{2}{4em}{60 mins}& RMSE & 11.60\% & 10.46\% &  9.42\% &  9.44\% &  \textbf{9.29\%} \\
                                  & MAE  &  7.91\% &  7.92\% &  \textbf{6.60\%} &  6.81\% &  \textbf{6.60\%} \\
    \hline
        \multirow{2}{4em}{120 mins}& RMSE & 15.18\% & 11.86\% & 11.93\% & 11.77\% & \textbf{11.55\%} \\
                                   & MAE  & 11.03\% &  8.72\% &  9.07\% &  8.55\% & \textbf{8.41\%} \\
    \hline
	\multirow{2}{4em}{180 mins}& RMSE & 17.84\% & 12.83\% & 12.94\% & 13.20\% & \textbf{12.26\%} \\
                                   & MAE  & 13.10\% & 10.13\% & 10.00\% &  9.84\% &  \textbf{9.15\%} \\
    \hline
   	\multirow{2}{4em}{240 mins}& RMSE & 20.32\% & 13.16\% & 13.02\% & 13.75\% & \textbf{12.52\%} \\
                                   & MAE  & 14.95\% & 10.16\% & 10.14\% & 10.08\% & \textbf{9.53\%} \\
\end{tabular}
\small
\caption{Normalised prediction error over different time horizons, with best results highlighted in bold. }
\end{table}

%% [{\bf{TODO}} is there anything that you could say comparing to the results already achieved by bigger models, for example from OCF, even if those are not exactly the same and validated on a different dataset??]
%% Yes - OCF have a normalised MAE over eight hours of about 6% 
%% https://drive.google.com/file/d/1sDKZ8WEJlTNa5oyonbNl2xGyZ7GLXKtQ/view

%%[{\bf{TODO}} Actually the PV only CNN seems to be more accurate than the Conv3D sat and pv.]
%% It seems to depend a lot on the error metric, and I think it's maybe quite noisy

%% [{\bf{TODO - Ben response in source text}} Why is RMSE and MAE in percentages, did you do any transformations to have them like that or is it a typo?]
%% The input and output is standardised 0-1 and the errors are the raw values, so if I got an error of 0.1316, I put it as 13.16% as this is what I saw in a similar ML dissertation. Was that not the right thing to do?

\subsection{Carbon emissions}

This section estimates the energy requirements of our models. 
The carbon costs are calculated through the time taken multiplied by estimates of the carbon intensity of time using a GPU.
Each of the models was run on a Tesla T4 GPU available through Google Colab, assumed to be in Northern Europe, the region closest to our location, which has a carbon efficiency of 0.21 kgCO$_2$eq/kWh. Energy usage simulations were conducted using the \href{https://mlco2.github.io/impact#compute}{Machine Learning Impact calculator} \cite{lacoste2019quantifying}. Table \ref{emissions-table} uses the computational time required to train and make inferences over a two-hour prediction window and compare the ConvLSTM and 3D CNN. It is assumed that the predictions are generated for each of the 330 UK grid supply points, and are generated every hour, from 06:00 to 20:00, and for a whole year, giving 330 * (20-6) * 365 = 1686,300 forecasts per year.
%Table \ref{emissions-table} shows the carbon costs associated with training these models.  

\begin{table}[H]
\label{emissions-table}
\scriptsize
\begin{center}
    \begin{tabular}{ p{10cm}  c  c }
        & \textbf{Conv3D} & \textbf{ConvLSTM}\\ \hline
        Time to train model (s) & 916.39 & 673.11 \\
        Time for inference, one forecast (s) & 0.16 & 2.21 \\
        Implies: Total time for year (training + 1686,300 * inferences) (hrs) & 75 & 1024 \\
        \hline
        \textbf{Emissions generated (tonnes $CO_{2}$-equiv) from \cite{lacoste2019quantifying}} & \textbf{0.0108} & \textbf{0.152} \\
        \hline
    \end{tabular}
    \caption{Estimates of emissions generated through training and deploying Conv3D and ConvLSTM}. 
    \small
\end{center}
\end{table}

The carbon reductions through increased model accuracy are harder to quantify. In 2020, the UK energy grid was estimated to produce 51 million tonnes of $C0_{2}$-equivalent emissions \cite{2020_uk_ghg_emissions}. Currently, 2\% of UK energy currently comes from solar PV, which could rise to 7\% by 2050 \cite{natgrideso_future_scenarios}. %A rough estimate could be that 1\% of energy supply is associated with 1 million tonnes of $C0_{2}$-equivalent emissions, though this will likely decrease over time.  
Emissions of $C0_{2}$ per gigawatt hour of electricity over the lifecycle of solar plant is 5 tonnes \cite{ritchie2021safest}, where
gigawatt hour is the annual electricity consumption of 150 Europeans. Based on this data, and assuming
67.22 million people in UK and a similar electrical consumption than Europeans, a 2\% of solar energy would generate annually 45,000 tonnes of emissions, which is 2450 times less than keeping gas-reserves online for generating the same electricity \cite{ritchie2021safest}.  Even if we assume that the models in Table \ref{emissions-table} are 0.1\% better at estimating solar output than the current standard of energy grid operators (which would be a lower bound as in our analysis these models are still much better than persistence, often a baseline followed for physical processes), and that in this 0.05\% of cases we may be able to turn off gas-reserves, we may be looking at a reduction of around 5500 tonnes of $C0_{2}$ annually.
%It is not possible to give precise benefits of the128 difference between ConvLSTM and Conv3D because the relationship between forecast accuracy and129 emissions averted is likely to be highly complex. 
%But given the amount of related C02-equivalent130 emissions, a small fractional improvement in forecast accuracy could be associated with thousands of tonnes C02-equivalent emissions per year within the UK alone. 
Given the costs of training the two models and generating the
forecasts for a year are between than 1/10th and 1/6th of a tonne of $C0_{2}$-equivalent, the benefits far outweigh the costs. This puts in perspective the reduction that could be possible if we were to deploy these low-resource models at large scale. 

\subsection{Analysing predictions and activations}
The task addressed by this paper was predicting a sequence of PV values, and was evaluated by performance on rMSE and MAE. But are the sequences of predictions realistic? Appendix \ref{prediction_error_analyis} examines the predictions given by the best model, the ConvLSTM, over a one-hour time horizon by overlaying predicted values an hour ahead with observed values. The predicted values generally trend up and down correctly suggesting the model has learned the dynamics of the sequence. However, this section also shows that the model consistently under-predicts PV yield across the year, most strongly in early afternoon in summer. This could be caused by the model having insufficient examples of sunny weather, long periods of low yield, and by the use of a symmetric loss function. Further work could explore whether loss functions which more heavily penalise under-predictions improve results.   

A frequent criticism of neural networks is that it is hard to interpret how they determine their results. Appendix \ref{interpretability} examines the activations in a simpler single-period neural network, to understand what drives the results, using an filter visualisation approach inspired by Chollet \cite{deep_learning_chollet}. This analysis suggests the model might be learning to recognise the contrast between land and sea. These activations would be associated with higher PV readings, as high contrast would mean a bright and clear day. But the model might not be serving our intended purpose of learning cloud dynamics. As a result, the model might perform worse on satellite areas with no coastline. Further work could examine the activations of the sequence models explored in the rest of this paper. 

\section{Conclusions}
%Why might the ConvLSTMs outperform CNNs over longer time scales? One reason could be that weather predictions images include both constant features and temporary aspects. For example, an important and constant feature might be a darker image overall, such as in the evening; and temporary aspect could be transient cloud pattern which might might build to provide cover, or might dissolve, leaving a clearer sky. The use of updated states in ConvLSTMs might provide further capacity to learn which features are constant and which ones are transient. This information might be present in satellite images but not in past PV data only. 

This paper includes an estimate of carbon emissions for training and running the inference of low-resource shallow neural architectures for solar photovoltaic nowcasting for a year. 
We attempt to estimate as well the emissions that would be averted by having a better integration of solar in the energy grid, showing that even under very pessimistic assumptions there is an important benefit in terms of total carbon footprint averted from integrating solar nowcasting machine learning models into the grid. Our analysis focuses on low-resource models that could be integrated into emerging economies, where energy requirements may be growing quickly. Further analyses are needed to understand the benefit that larger and deeper models could bring. 

The paper also examined the prediction errors and activations of a similar but simpler neural network. This suggested the model might be influenced by an unbalanced dataset, and that an augmented dataset or a different loss function might improve results. Analysis of the activations suggested the model might be learning to recognise contrast between land and sea, and further research could examine performance on satellite images with no coastline. 

A more in depth analysis of the current strategies implemented by grid operators used to meet energy demands in case of disruption to the renewable solar supply would be needed to better put in perspective the climate change mitigation potential of nowcasting models. 
%We provide a rough estimate of emissions that shows that even these models through better forecasts, and emissions generated through training and deploying models like this for one year. Depending on assumptions, there is at least a 100x benefit from better predictions, and we showed that relatively simple models beat persistence. So this seems like a potentially high-leverage way to reduce emissions. 
%Rich Sutton \cite{bitter_lesson} argued that the `bitter lesson' of AI research is that general methods that leverage computation are ultimately the most effective and accurate, outperforming handcrafted methods. But in this case, time is an important dimension and our explicit representation leads to better results. 

%\section{Acknowledgements}

%%%%%%%%%%%%%%%%%%%%%%%% APPENDIX %%%%%%%%%%%%%%%%%%%%%%%%%%%%%%%%%%%%%%%%%
\newpage
\appendix

\section{Analysing the predictions}\label{prediction_error_analyis}

This section explores the predictions for the best model, the ConvLSTM model, over a one-hour time horizon. It begins by comparing the predictions against true values for three randomly chosen days in the test set. 

\begin{figure}[ht]
\begin{center}
\includegraphics[width=12cm]{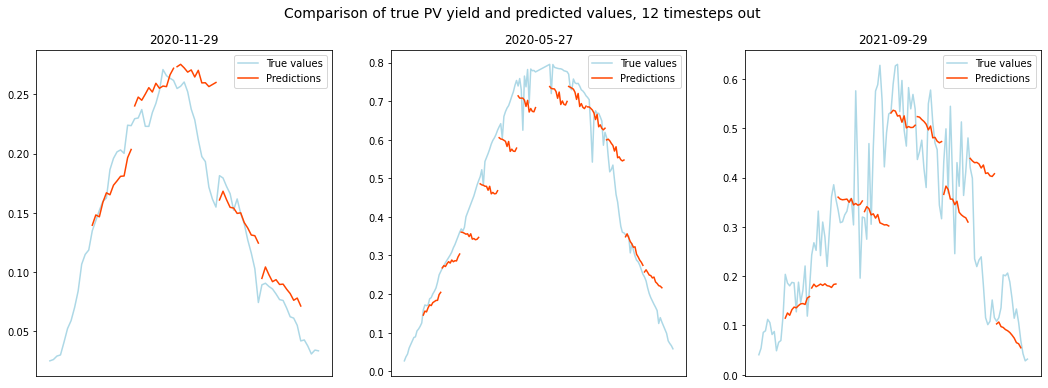}
\caption{Predicted and observed PV yield across three randomly chosen days in the test set. Predictions an hour ahead are in orange and observed values in light blue. Note the variation in y values (solar yield) across days. The model tends to under-predict sunny and cloudy days. The day on the left is a darker winter day, the middle day is sunny from summer, and the right day is a cloudy day in autumn.}
\end{center}
\end{figure}

Interestingly, it appears that the ConvLSTM generally predicts the overall gradient of the prediction correctly - if the prediction window is getting brighter or darker. However, in the middle day shown above, the model consistently predicts that the PV yield will decrease. The middle day has PV readings up to 75\% of the two-year maximum, and is much brighter than the other two examples which peak at around 30\% and 60\%. This suggests that the model does not deal well with sunnier days, and might under-predict on those days. 

The plot below shows the total relative prediction errors per mean prediction error over the sequence, across a two-year period. Since there are 12 predicted values with a range of [0,1], the largest possible magnitude of error is 12. The models optimise for mean square error, which is not sensitive to the sign of the error. This plot shows the direction of the error. 

\begin{figure}[ht]
\begin{center}
\includegraphics[width=12cm]{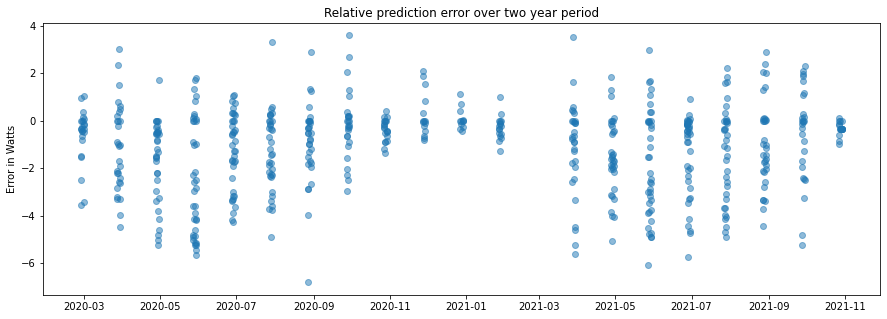}
\caption{Plot of relative prediction error over two years - mean error over sequence. Overall the model tends to have a negative error - unpredicts PV yield. This is more extreme in summer, with smaller errors in winter. }
\end{center}
\end{figure}

It appears that the model tends to have the errors with the largest magnitude in summer, lower magnitude errors in November, December, and January. Since the test set days are three days chosen from the end of each month, the values appear in lines which span the test set ranges.

\section{Interpretability}\label{interpretability}

A frequent criticism of neural networks is that it is hard to interpret how they determine their results. For simplicity, a simpler neural network was trained to infer a single PV reading from a single image. The model was trained for 10 epochs on a randomly selected sample of 1000 satellite images, and it achieved a test set performance of a 2\% mean square error. Below are the activations for three images taken at different times with different characteristics. 

\textbf{Activations for Case A - Bright conditions, high PV}

\begin{figure}[ht!]
  \centering
  \begin{minipage}[b]{0.3\textwidth}
    \includegraphics[width=\textwidth]{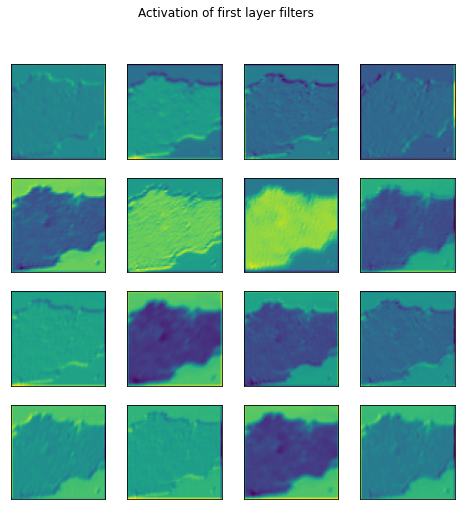}
  \end{minipage}
  \hfill
  \begin{minipage}[b]{0.3\textwidth}
    \includegraphics[width=\textwidth]{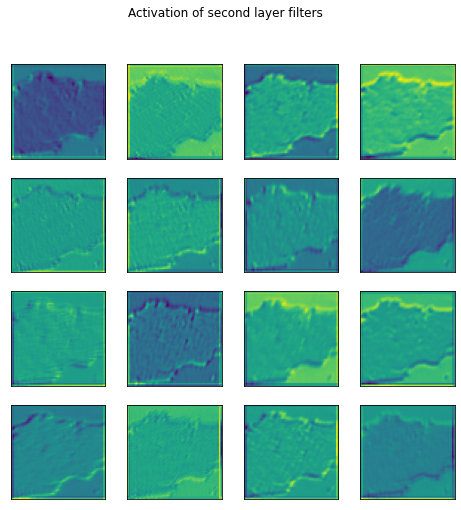}
  \end{minipage}
  \hfill
    \begin{minipage}[b]{0.3\textwidth}
    \includegraphics[width=\textwidth]{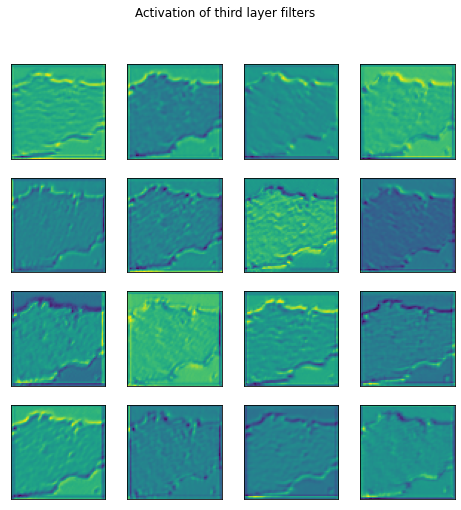}
  \end{minipage}
\end{figure}

\noindent
\textbf{Activations for Case B - Dark conditions, some cloud}

\begin{figure}[ht!]
  \centering
  \begin{minipage}[b]{0.3\textwidth}
    \includegraphics[width=\textwidth]{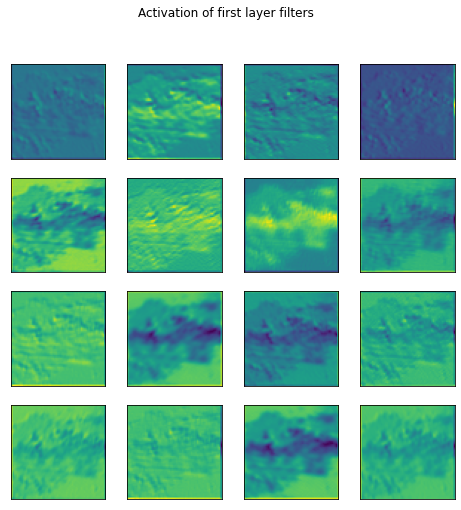}
  \end{minipage}
  \hfill
  \begin{minipage}[b]{0.3\textwidth}
    \includegraphics[width=\textwidth]{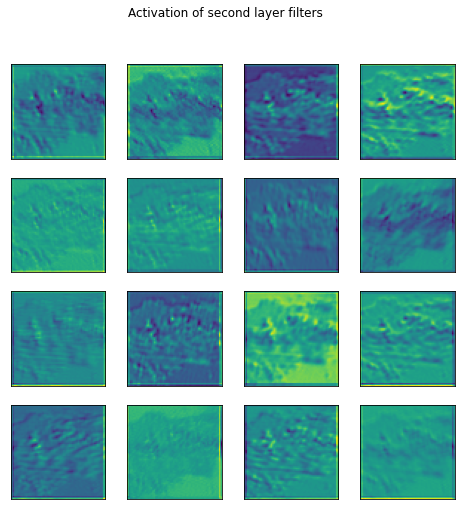}
  \end{minipage}
  \hfill
    \begin{minipage}[b]{0.3\textwidth}
    \includegraphics[width=\textwidth]{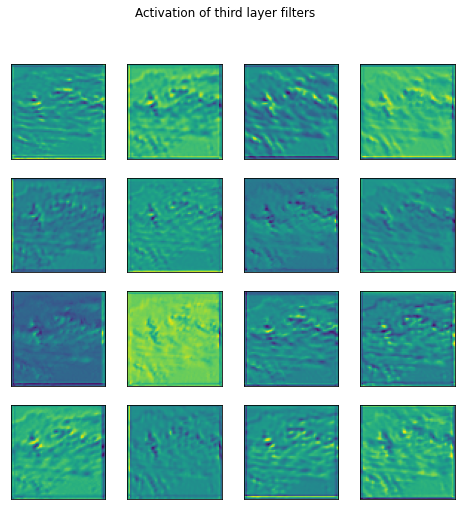}
  \end{minipage}
\end{figure}

\noindent
\textbf{Activations for Case C - Bright conditions, thick cloud}

\begin{figure}[ht!]
  \centering
  \begin{minipage}[b]{0.3\textwidth}
    \includegraphics[width=\textwidth]{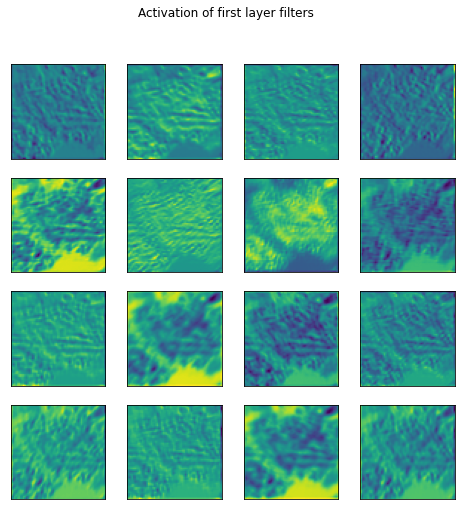}
  \end{minipage}
  \hfill
  \begin{minipage}[b]{0.3\textwidth}
    \includegraphics[width=\textwidth]{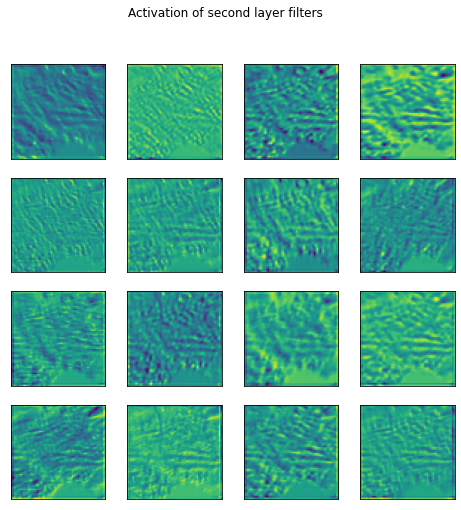}
  \end{minipage}
  \hfill
    \begin{minipage}[b]{0.3\textwidth}
    \includegraphics[width=\textwidth]{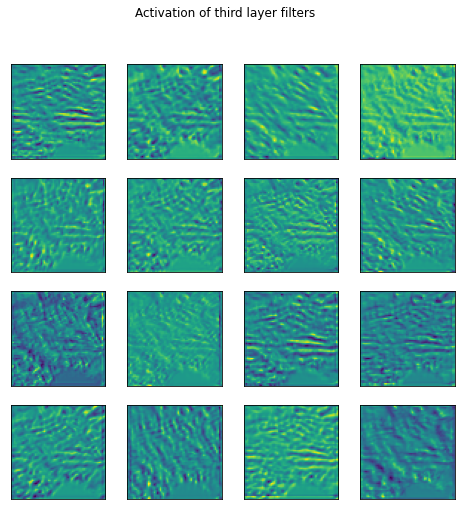}
  \end{minipage}
\end{figure}

On the first layer on the left we see bright activations for the sea and the land, as expected since the first layer of a CNN often learns textures. The second layer shows high activation around the coast line and the sea, and the third layer on the right seems to be mostly highly activated around the coast line. 

From these examples it seems that this network is learning to recognise contrast between land and sea. Intuitively, a high contrast will mean a bright sky with no clouds. This will be when the PV is highest, and so the activations will be associated with better predictions. 

But in this case the model might not have learnt any cloud features. This might also be true for our main models - are they also just recognising contrast? A useful area of further research would be examining the activations in the more complex, time series models used in the rest of the paper.

% \section{Assumptions used in carbon cost estimates}\label{cost-estimates}

\printbibliography

@online{eumetsat_dataset,
    title = "EUMETSAT Dataset",
    URL = "https://huggingface.co/datasets/openclimatefix",
    author = "Open Climate Fix",
    year = "2022"}

@online{2020_uk_ghg_emissions,
    title = "2020 UK greenhouse gas emissions",
    year = "2021",
    author = "UK Office for National Statistics",
    URL = "https://www.gov.uk/government/statistics/provisional-uk-greenhouse-gas-emissions-national-statistics-2020"}

@article{systematic-reporting-footprints,
author = {Henderson, Peter and Hu, Jieru and Romoff, Joshua and Brunskill, Emma and Jurafsky, Dan and Pineau, Joelle},
title = {Towards the Systematic Reporting of the Energy and Carbon Footprints of Machine Learning},
year = {2020},
issue_date = {January 2020},
publisher = {JMLR.org},
volume = {21},
number = {1},
issn = {1532-4435},
journal = {J. Mach. Learn. Res.},
month = {jan},
articleno = {248},
numpages = {43}
}

@incollection{GAO201535,
title = {Chapter 2 - Applications of ESS in Renewable Energy Microgrids},
editor = {David Wenzhong Gao},
booktitle = {Energy Storage for Sustainable Microgrid},
publisher = {Academic Press},
address = {Oxford},
pages = {35-77},
year = {2015},
isbn = {978-0-12-803374-6},
doi = {https://doi.org/10.1016/B978-0-12-803374-6.00002-0},
url = {https://www.sciencedirect.com/science/article/pii/B9780128033746000020},
author = {David Wenzhong Gao}
}

@online{natgrideso_future_scenarios, 
    title = "Future Energy Scenarios: July 2022", 
    year = "2022", 
    author = "Natioal Grid ESO",
    URL = "https://www.nationalgrideso.com/future-energy/future-energy-scenarios"}

@article{ritchie2021safest,
  title={What are the safest and cleanest sources of energy},
  author={Ritchie, Hannah},
  journal={Our World Data. https://ourworldindata. org/safest-sources-of-energy. Accessed},
  volume={22},
  year={2021}
}

@inproceedings{ConvLSTM,
 author = {Shi, Xingjian and Chen, Zhourong and Wang, Hao and Yeung, Dit-Yan and Wong, Wai-kin and WOO, Wang-chun},
 booktitle = {Advances in Neural Information Processing Systems},
 editor = {C. Cortes and N. Lawrence and D. Lee and M. Sugiyama and R. Garnett},
 pages = {},
 publisher = {Curran Associates, Inc.},
 title = {Convolutional LSTM Network: A Machine Learning Approach for Precipitation Nowcasting},
 url = {https://proceedings.neurips.cc/paper/2015/file/07563a3fe3bbe7e3ba84431ad9d055af-Paper.pdf},
 volume = {28},
 year = {2015}
}

@inproceedings{boucher-clouds,
    author = "Olivier Boucher and David Ransall",
    title = "Clouds and Aersols",
    booktitle = "Climate Change 2013: The Physical Science Basis. Contribution of Working Group I. to the 5th Assessment Report of the Intergovernmental Panel on Climate",
    pages = "571-657",
    year = "2013",
    publisher = "Cambridge University Press"}

@book{deep_learning_chollet,
    title = "Deep Learning with Python (Second Edition)",
    year = "2021",
    author = "Francois Chollet", 
    publisher = "Manning"}

@online{cnns-vs-lstms, 
    title="An Empirical Evaluation of Generic Convolutional and Recurrent Networks for Sequence Modeling", 
    author = "Shaojie Bai and J. Zico Kolter and Vladlen Koltun",
    URL = "https://arxiv.org/abs/1803.01271"}

@online{natgrideso_road_zero_carbon,
    title = "The Road to Zero Carbon",
    year = "2022",
    author = "National Grid ESO",
    URL = "https://www.nationalgrideso.com/future-energy/our-progress/road-zero-carbon/report"
}

@article{lacoste2019quantifying,
  title={Quantifying the Carbon Emissions of Machine Learning},
  author={Sasha Luccioni and Victor Schmidt and Alexandre Lacoste and Thomas Dandres},
  booktitle={NeurIPS 2019 Workshop on Tackling Climate Change with Machine Learning},
  url={https://www.climatechange.ai/papers/neurips2019/22},
  year={2019}
}

@article{royal-soc-can-beat,
    title = "Can deep learning beat numerical weather prediction?",
    author = "Martin Schultz and Clara Betancourt and Bing Gong and Felix Kleinert and Michael Langguth and Lukas Leufen and Amirpasha Mozaffari and Scarlet Stadtler",
    year = {2021},
    month = {02},
    title = {Can deep learning beat numerical weather prediction?},
    volume = {379},
    journal = {Philosophical Transactions of The Royal Society A},
    doi = {https://doi.org/10.1098/rsta.2020.0097}
}

\end{document}